\pdfoutput=1

\documentclass[11pt]{article}

\usepackage[final]{acl}

\usepackage{times}
\usepackage{latexsym}
\usepackage{amsmath}
\usepackage{amssymb}
\usepackage{breqn}
\usepackage{amssymb}
\usepackage{color, soul}
\usepackage{hyperref}
\usepackage{float}
\usepackage{arydshln}
\usepackage{multirow}
\usepackage{multicol}
\usepackage{xcolor,pifont}
\usepackage{tcolorbox}
\usepackage{fvextra} 
\newcommand*\colourcheck[1]{%
  \expandafter\newcommand\csname #1check\endcsname{\textcolor{#1}{\ding{52}}}%
}
\newcommand*\colourtimes[1]{%
  \expandafter\newcommand\csname #1times\endcsname{\textcolor{#1}{\ding{55}}}%
}
\colourcheck{green}
\colourtimes{red}
\colourcheck{black}
\colourtimes{black}

\usepackage[T1]{fontenc}

\usepackage[utf8]{inputenc}

\usepackage{microtype}

\usepackage{inconsolata}

\usepackage{graphicx}

\usepackage{amsfonts}

\title{\textbf{\texttt{TrueBrief}}: Faithful Summarization through Small Language Models}

\author{Kumud Lakara \\
  \texttt{kumud.lakara@jpmorgan.com} \\\And
  Ruibo Shi \\
  \texttt{ruibo.shi@jpmchase.com} \\\And
  Fran Silavong \\
  \texttt{fran.silavong@jpmchase.com}}
\newcommand\blfootnote[1]{%
  \begingroup
  \renewcommand\thefootnote{}\footnote{#1}%
  \addtocounter{footnote}{-1}%
  \endgroup
}
\begin{document}
\maketitle
\begin{abstract}
    Large language models (LLMs) have exhibited remarkable proficiency in generating high-quality text; however, their propensity for producing hallucinations poses a significant challenge for their deployment in security-critical domains. In this work, we present \textbf{\texttt{TrueBrief}}, an end-to-end framework specifically designed to enhance the faithfulness of small LLMs (SLMs) primarily for the task of text summarization through a preference-optimization paradigm. Central to our framework is a data generation module that facilitates controlled hallucination injection to generate synthetic preference data. 
    Our work provides insights into the impact of data quality and model size on preference-based optimization, highlighting the conditions under which these methods are most effective.
\end{abstract}
\blfootnote{This paper was prepared for informational purposes by the Machine Learning Center of Excellence group of JPMorgan Chase \& Co. and its affiliates (``JP Morgan''),and is not a product of the Research Department of JP Morgan. JP Morgan makes no representation and warranty whatsoever and disclaims all liability, for the completeness, accuracy or reliability of the information contained herein. This document is not intended as investment research or investment advice, or a recommendation, offer or solicitation for the purchase or sale of any security, financial instrument, financial product or service, or to be used in any way for evaluating the merits of participating in any transaction, and shall not constitute a solicitation under any jurisdiction or to any person, if such solicitation under such jurisdiction or to such person would be unlawful.}
\begin{figure*}[h]
    \centering
    \includegraphics[width=\linewidth]{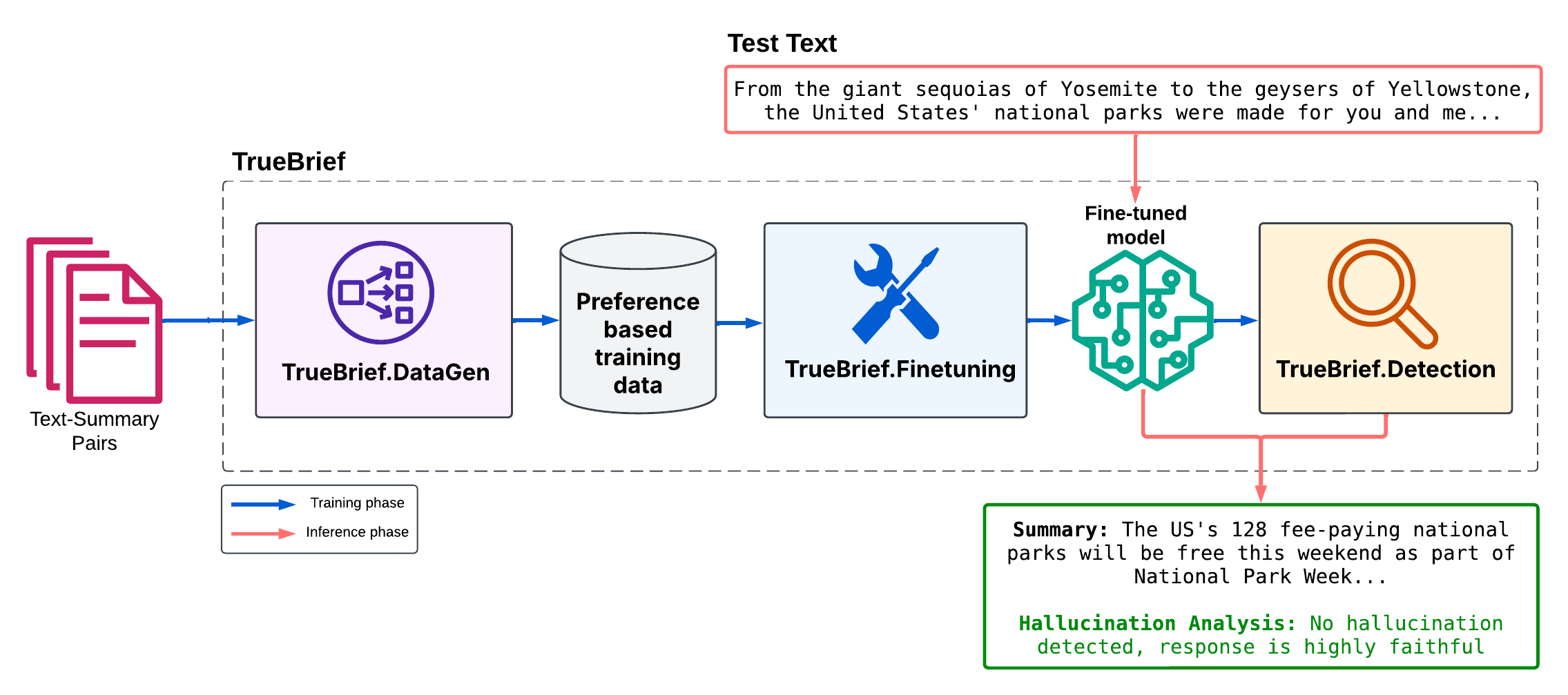}
    \caption{\textbf{\texttt{TrueBrief} Components Overview}: The main pipeline leverages the \texttt{data generation} module, which is pivotal to the system's performance due to its novel hallucination injection component for creating a preference-based dataset. This dataset is used to finetune a given model via the \texttt{finetuning} module. The resulting finetuned model is employed for response generation and integrated into the \texttt{detection} module, which utilizes a white-box approach to effectively detect hallucinations.}
    \label{fig:main_figure}
\end{figure*}

\section{Introduction}
Recent works have shown that small LLMs (SLMs), with finetuning, are capable of competing with and in certain scenarios outperforming general purpose larger models~\cite{niu-etal-2024-ragtruth}. SLMs such as the Qwen2.5~\cite{qwen2.5} and Phi-3~\cite{abdin2024phi3technicalreporthighly} families of models have recently exhibited remarkable capabilities across a wide range of natural language processing (NLP) tasks including summarization, question-answering and machine translation~\cite{open_llm_leaderboard}. In addition, when compared to larger models, SLMs are more cost-efficient, have a lower environmental impact and are relatively faster and cheaper to deploy~\cite{slms_cheaper, wang2024comprehensivesurveysmalllanguage, projected-language-models}. They are also particularly suited to situations where access to larger models through APIs is not possible due to regulatory or security reasons. However, like all language models, the usability of SLMs is hampered by their tendency to hallucinate~\cite{xu2024hallucinationinevitableinnatelimitation, llm_hallucination_survey, calibrated_hallucination}.
To this end, in this work we focus on enhancing faithfulness of SLMs by proposing an end-to-end framework that adopts a dual-faceted approach to \emph{reduce} and \emph{detect} hallucinations in model generated responses.


We introduce \textbf{\texttt{TrueBrief}}, a framework for faithful summarization using SLMs. We propose using a preference tuning objective, namely, Direct Preference Optimization (DPO)~\cite{rafailov2024directpreferenceoptimizationlanguage} along with a novel synthetic data generation process that creates high-quality preference pairs. Using this set-up we specifically analyse the effect of data quality and model size on preference-based optimization, establishing optimal conditions for its use. We also extend existing white-box hallucination detection methods and integrate them into our pipeline to create an end-to-end framework complete with summarization and hallucination detection capabilities. Through the detection component of our pipeline we show that SLMs with as few as 0.5 billion parameters can achieve reasonable performance in detecting hallucination in responses generated by significantly larger models such as GPT-3.5-turbo~\cite{openai_gpt35_turbo} and Llama-2-70B~\cite{touvron2023llama2openfoundation}. 
Our findings reveal that internal model dynamics of language models, regardless of scale, share underlying commonalities which can be effectively leveraged for hallucination detection. This insight is underscored by our ability to utilize SLMs to detect hallucinations in outputs from larger, distinct models, demonstrating a promising cross-model applicability in this domain.

The main contributions of this work are:
\begin{itemize}
    \item An end-to-end framework to enhance faithfulness of SLMs through integrated components for data generation, finetuning and hallucination detection.
    \item A synthetic data generation method for creating preference-based dataset via controlled hallucination injections.
    \item Novel insights into how data quality and model size affect preference-based optimization for faithfulness.
    \item An extension of existing low latency white-box hallucination detection methods that leverages internal model dynamics to facilitate real-time detection without reliance on external APIs or LLM calls.
\end{itemize}
While the focus of this work is the summarization task, it is feasible to extend our proposed method to other RAG tasks such as Question-Answering and Machine Translation.

\section{Related Work}
\textbf{Hallucination Reduction: }~\cite{si2023promptinggpt3reliable} use prompt engineering to generate more reliable responses from GPT-3~\cite{gpt3} by explicitly instructing the model to not make assumptions and be fair. Retrieval augmented generation (RAG) has also been actively explored to reduce hallucinations in generated responses~\cite{semnani-etal-2023-wikichat, shuster-etal-2021-retrieval-augmentation, varshney2023stitchtimesavesnine}. Specifically,~\cite{peng2023checkfactstryagain} present LLM-Augmenter which uses a series of plug-and-play modules to use external knowledge to guide an AI agent to generate more faithful responses.~\cite{gao-etal-2023-rarr} propose RARR where disagreement between the generated text and retrieved evidence is used to fix hallucination in generated text. 
~\cite{dhuliawala-etal-2024-chain}, propose the Chain-of-Verification method where a model refines its responses through a question-answering process.~\cite{ji-etal-2023-towards} present a self-reflection methodology that involves knowledge acquisition and answer generation to enhance the factuality, consistency and entailment of the generated responses.~\cite{jones2023teachinglanguagemodelshallucinate} and~\cite{cheng-etal-2023-uprise} use prompt-tuning to reduce hallucination in the generated model responses. Further related to our work, a series of research focuses on using SFT~\cite{yang2024alignmenthonesty, zhang-etal-2024-r} and Reinforcement Learning (RL)~\cite{sun2023aligninglargemultimodalmodels, mesgar2021improvingfactualconsistencyresponse, roit-etal-2023-factually} for hallucination reduction. In particular,~\cite{hu2024mitigatinglargelanguagemodel} introduce faithful finetuning which is a multi-task training method consisting of separate sub objectives to train the model to effectively retrieve internal knowledge and generate factually grounded responses.~\cite{tian2023finetuninglanguagemodelsfactuality} use Direct Preference Optimization~\cite{rafailov2024directpreferenceoptimizationlanguage} to align model behaviour for factuality.~\cite{kang2024unfamiliarfinetuningexamplescontrol} propose a method to improve the factuality of generated responses using RL with conservative reward models.~\cite{shi-etal-2024-trusting} show contrastive decoding to be effective in mitigating hallucination when generating context based responses.~\cite{wang2024directjudgementpreferenceoptimization} combine SFT and DPO objectives, however, they primarily focus on the task of judgment capabilities of LLMs, and their work is limited to the DPO~\cite{rafailov2024directpreferenceoptimizationlanguage} method for preference alignment. RAG-HAT~\cite{raghat} is the most related to our work and involves training a detection model to output detection labels and descriptions which are used to correct hallucinations in the ground truth data to create a preference-based dataset. 

\textbf{Hallucination Detection:}~\cite{belyi2024lunaevaluationfoundationmodel} finetune a DeBERTA-large encoder for hallucination detection in RAG settings. ~\cite{manakul2023selfcheckgptzeroresourceblackboxhallucination} use the consistency between stochastically sampled responses for a given input to decide if the generated response contains hallucination or not.~\cite{cohen2023lmvslmdetecting} propose a factuality evaluation framework based on cross-examination through multi-turn interaction between LLMs. ~\cite{kossen2024semanticentropyprobesrobust, ch-wang-etal-2024-androids} train simple linear classifiers called probes on model hidden states to capture intrinsic features specific to responses containing hallucinations. ~\cite{cheng2024smallagentrockempowering, chern2023factoolfactualitydetectiongenerative,min2023factscorefinegrainedatomicevaluation} use external retrieval to identify (factual) inconsistencies in the generated responses which are then used to detect hallucination in generated responses. Finally, ~\cite{manakul2023selfcheckgptzeroresourceblackboxhallucination, kossen2024semanticentropyprobesrobust, varshney2023stitchtimesavesnine} adopt a more statistical approach and make use of uncertainty based metrics such as entropy over the model's output distribution to detect hallucination.

\textbf{DPO:}~\cite{rafailov2024directpreferenceoptimizationlanguage} propose the Direct Preference Optimization (DPO) algorithm as a potential alternative to RLHF. DPO encourages the model to favour responses aligned with human preferences (hallucination-free responses, in our case) over other dis-preferred responses (hallucinated responses, in our case). Each data sample required for preference based finetuning consists of an input prompt $x$, a preferred response $y_w$ and a rejected response $y_l$; which gives us, $\mathcal{D}^{dpo}=\{x^{(i)},y^{(i)}_w, y^{(i)}_l\}_{i=1}^N$. Beyond preference pairs, ~\cite{rafailov2024directpreferenceoptimizationlanguage} also introduce the Plackett-Luce ranking model~\cite{plackett, luce1959individual}. This version is compatible with frameworks which have access to multiple (>2) ranked answers. While ~\cite{rafailov2024directpreferenceoptimizationlanguage} demonstrate that DPO can be used to align models with human behaviour, they do not experiment with model faithfulness. Work exploring the optimal settings to use DPO is also limited. Our work, specifically focuses on leveraging DPO in order to enhance model faithfulness while extensively exploring the optimal conditions for using DPO in relation to model size and data quality.

\section{\textbf{\texttt{TrueBrief}}}
\textbf{\texttt{TrueBrief}} is an end-to-end training and inference framework aimed at enhancing model faithfulness through three integrated components: \textbf{\texttt{TrueBrief.DataGen}},\textbf{\texttt{TrueBrief.Finetuning}} and \textbf{\texttt{TrueBrief.Detection}} which work together to generate faithful summaries using SLMs. 
Figure~\ref{fig:main_figure} presents a detailed overview of how these components work in tandem.
\subsection{\textbf{\texttt{TrueBrief.DataGen}}}
\label{subsec: datagen}
 \begin{table}[h]
\centering
\begin{tabular}{llll}
\hline
\textbf{Intrinsic} & \textbf{Extrinsic} & \textbf{Extrinsic} & $\mathbf{F_{score}}$ \\
   & \textbf{(low)} & \textbf{(high)} &  \\
\hline
\redtimes & \redtimes & \redtimes & 0.73 \\
\greencheck & \redtimes & \redtimes & 0.71 \\
\greencheck & \redtimes & \greencheck & 0.71 \\
\greencheck & \greencheck & \redtimes & 0.77 \\
\greencheck & \greencheck & \greencheck & 0.74 \\
\hline
\end{tabular}
\caption{\label{tab: data_and_hallucination}
  Effect of data quality on faithfulness for Qwen2.5-0.5B-Instruct.
}
\end{table}
The \textbf{\texttt{TrueBrief.DataGen}} module facilitates controlled hallucination injection into groundtruth samples (chosen responses) to generate the corresponding rejected responses. The data generation process involves a two-fold hallucination injection strategy, focusing on \emph{intrinsic} and \emph{extrinsic} hallucination, where the intrinsic involves augmentation of factual entities and for extrinsic, we randomly paraphrase and introduce ungrounded information in one or more original sentences for low and high hallucination levels. Including both types is crucial for a comprehensive evaluation of the model's ability to detect and mitigate different forms of hallucinations. Intrinsic hallucination assesses the model’s sensitivity to factual accuracy within the provided context, whereas extrinsic evaluation measures the model’s ability to distinguish relevant information from irrelevant or misleading external content.

The data generation process involves factually augmenting the ground truth summary to inject factual hallucination into the data. We extract named entities using SpaCy~\cite{spacy2} and then replace the extracted entities using an LLM with false values. This is followed by systematic paraphrasing of parts of the factually augmented response. We randomly select between at least one sentence to 100\% of the input summary to paraphrase. This process ensures the final hallucinated response does not deviate too far away from the ground truth response in terms of lexical and semantic style. Comprehensive details of the data generation component and the controlled hallucination injection method are provided in Appendix~\ref{app:sys_components}. 

The hallucination injection is executed in a controlled manner to ensure that the semantic and lexical style of the response remains consistent with the ground truth response. We find this crucial to the data generation process, and in the following sections, show that uncontrolled hallucination injection can reduce the effectiveness of the preference-optimization paradigm. This can lead the model, during finetuning, to focus on the semantic style of the chosen response rather than adhering to the intended behaviour of enhanced faithfulness. 

\subsection{\texttt{TrueBrief.Finetuning}}
We employ DPO~\cite{rafailov2024directpreferenceoptimizationlanguage} as the core preference based optimization objective to train the models in this work. We use the data generated by \textbf{\texttt{TrueBrief.DataGen}}(\autoref{subsec: datagen}) to finetune models to enhance their faithfulness. Through preference-based optimization we are able to finetune a given language model for a downstream task while encouraging model faithfulness with respect to the reference source. 
Specifically, we extend DPO to the task of aligning model behaviour to be highly faithful, where for a given input ($x$), we designate the chosen response ($y_w$) as one free from hallucinations and the rejected response ($y_l$) as the synthetically generated hallucinated response. Intuitively, we aim to reinforce model behaviour that penalizes various forms of hallucination, enabling the model to implicitly learn to generate more faithful responses. 

We also investigate whether incorporating multiple rejected responses (as opposed to only one) would be advantageous for preference-based optimization. We specifically explore whether \textit{``less is more''} or \textit{``more is more''} in the context of preference-based optimization. To this end, we propose two variations of DPO by reformulating the data setting to require multiple (>1) rejected responses for every chosen response. Rejected responses are generated by progressively increasing the level of extrinsic hallucination from paraphrasing just one (low) sentence to paraphrasing 50\% to 100\% (high) of the source text. We remain indifferent between the rejected responses, expressing preference solely for the chosen response, giving us an extended version of $\mathcal{D}^{dpo}$ in the following form:
\begin{equation}
    \mathcal{D}^{dpo}_{extended}=\{x^{(i)}, y_w^{(i)}, y_{l_1}^{(i)}, y_{l_2}^{(i)}, .. y_{l_{k-1}}^{(i)}\}_{i=1}^N 
\end{equation}
 With strict preference for $y_w$ over $y_{l_j}$ i.e. $y_w \succ y_{l_j} \forall j \in \{1,2,..,k-1\}$ (with $\succ$ indicating strict preference) and being indifferent between the rejected responses i.e $y_{l_1}\sim y_{l_2}...\sim y_{l_{k-1}}$ (with $\sim$ indicating mutual indifference). For the experiments presented in this work, we use 3 rejected response and hence set $k=4$. The presence of multiple rejected responses, renders the Bradley-Terry~\cite{bradley-terry} model, which forms the base for DPO, inadequate. Therefore for data settings where we have multiple rejected responses and there is no preference order between them, we propose variations to explore the impact of using multiple rejected responses versus just one: \\
 \begin{table*}[h]
\small
\centering
\begin{tabular}{llllllllllll}
\hline
\textbf{Model} & \textbf{Model} & \multicolumn{3}{c}{\textbf{Rouge Score}} & \textbf{BERT} & \multicolumn{4}{c}{\textbf{LLM-as-Judge}} & $\mathbf{F_{score}}$ &$\mathbf{B_{score}}$\\
\textbf{Size}& &\textbf{R-1} & \textbf{R-2} & \textbf{R-L} &\textbf{score} &\textbf{Comp.} & \textbf{Rele.} & \textbf{Flue.} & \textbf{Cohe.}&  & \\
\hline
\multirow{5}{*}{0.5B} & Baseline & 0.40 & 0.16 & 0.25 & 0.64 &2.57 & 3.18  &\textbf{3.92}&\textbf{3.60}&0.72& 0.62 \\
& SFT & 0.44 & 0.19& \textbf{0.29} &0.66 &2.54&3.15&3.91&3.52& 0.73&  0.62 \\
& DPO & 0.44 & 0.19 & 0.28 & 0.66&2.66&3.19&3.91&3.56&\textbf{0.77}&\textbf{0.65}\\
& Sep-DPO&0.45 &0.19&0.28&\textbf{0.67}&\textbf{2.69}&\textbf{3.21}&3.85&3.52&0.74&0.64\\
& Add-DPO & 0.43 & 0.17 & 0.27&0.65 &2.54&3.12&3.83&3.47&0.74&0.62\\
& PL-DPO&\textbf{0.45}&\textbf{0.20}&\textbf{0.29}&\textbf{0.67}&2.54&3.13&3.85&3.50&0.73&0.62\\
\hline
\multirow{5}{*}{1.5B} & Baseline & 0.41 & 0.16 & 0.25 &0.65&\textbf{3.22}&3.71&\textbf{4.45}&\textbf{4.30}&0.81&0.73\\
& SFT & 0.45 & 0.20 & 0.29 &0.67&3.15&3.63&4.23&4.02&0.84&0.73\\
& DPO & \textbf{0.46} & 0.20 & \textbf{0.30}&\textbf{0.68}&3.20&\textbf{3.81}&4.44&4.23&\textbf{0.86}&\textbf{0.75}\\
& Add-DPO & \textbf{0.46} & \textbf{0.21} & \textbf{0.30}&\textbf{0.68} &3.15&3.71&4.40&4.17&0.84&0.73\\
& PL-DPO & \textbf{0.46} & 0.20 & \textbf{0.30} &\textbf{0.68}&\textbf{3.22}&3.67&4.37&4.16&0.85&0.74\\
\hline
\multirow{5}{*}{3B}&Baseline & 0.47 & 0.21 & 0.31&0.68 
&3.30&\textbf{4.23}&4.68&4.53&0.91&0.79\\
& SFT & \textbf{0.52} & \textbf{0.26} & \textbf{0.35} &\textbf{0.71}&\textbf{3.52}&4.22&4.72&\textbf{4.59}&\textbf{0.93}&\textbf{0.82}\\
& DPO& 0.51& 0.25&0.34&0.70&3.43&4.09&4.71&4.48&\textbf{0.93}&\textbf{0.82}\\
& Add-DPO&0.51&0.24&0.33&0.70&3.46&4.16&\textbf{4.73}&4.54&\textbf{0.93}&0.81\\
& PL-DPO&0.51&\textbf{0.26}&0.34&0.70&3.51&4.05&4.68&4.51&0.92&0.81\\
\hline
\multirow{5}{*}{7B}&Baseline & 0.47 & 0.20 & 0.31 & 0.68&3.30&4.24&4.73&4.60&0.95&0.80\\
& SFT & \textbf{0.52} & \textbf{0.26} & \textbf{0.35} &\textbf{0.71}& 3.79&4.38&4.83&4.72&\textbf{0.96}&0.85\\
& DPO & \textbf{0.52}&\textbf{0.26}&\textbf{0.35}&\textbf{0.71}&\textbf{3.80}&\textbf{4.39}&\textbf{4.85}&\textbf{4.73}&\textbf{0.96}&\textbf{0.86}\\
\hline
\end{tabular}
\caption{\label{tab: finetuning_results}
Model response quality analysis for Qwen2.5 family of models. Comp, Rele, Flue, Cohe correspond to Compelteness, Relevance, Fluency and Coherence dimensions of the LLM as Judge evaluation. $F_{score}$ and $B_{score}$ refer to the faithfulness and balanced scores respectively as defined in \autoref{subsec:metrics}. A complete version of this table containing respective standard deviation values is presented in table~\ref{tab: full_finetuning_results} in Appendix~\ref{app:complete_finetuning_results}.
}
\end{table*}
\noindent
\textbf{Additive (Add)-DPO:} In order to accommodate $\mathcal{D}^{dpo}_{extended}$ we replace the term corresponding to the rejected responses in standard DPO with an average over all the rejected responses in the log space, resulting in the additive DPO (add-DPO) loss function.
\label{eqn: add-dpo}
\begin{dmath}
    \mathcal{L}_{\text{Add-DPO}}(\pi_\theta, \pi_{ref}) = 
    -\mathbb{E}_{(x,y_w,\{y_{l_i}\})\sim \mathcal{D}^{dpo}_{\text{extended}}} \bigg[\\
    \log\sigma\bigg(\beta\log\left(\frac{\pi_\theta(y_w|x)}{\pi_{ref}(y_w|x)}\right)
    -\frac{\beta}{k}\sum_{i=1}^{k-1}\log\left(\frac{\pi_\theta(y_{l_i}|x)}{\pi_{ref}(y_{l_i}|x)}\right)\bigg)\bigg]
\end{dmath}
Where $\beta$ is a parameter controlling the deviation of the language model (policy) $\pi_\theta$ from the base reference policy $\pi_{ref}$.

\textbf{Modified Plackett-Luce (PL)-DPO:}
 A key point of difference between the existing PL-DPO formulation as proposed by ~\cite{rafailov2024directpreferenceoptimizationlanguage} and our specific use case is the relative preference ranking of responses. While the PL preference model~\cite{plackett, luce1959individual} assumes a preference ranking between \textit{all} responses (a more complex data setting), we are mutually indifferent between responses which contain hallucination. Therefore, we modify the existing PL-DPO formulation and refer to it as $\mathcal{L}_{\text{PL-DPO}}$.
 \begin{dmath}
    \mathcal{L}_{PL-DPO} = \\
    -\mathbb{E}{(x, y_w, {y_{l_i}})\sim\mathcal{D}_{extended}^{DPO}} \left[\\
    -\log\left(
    \frac{ exp \left( \beta \log \frac{\pi_\theta(y_w|x)}{\pi_{\text{ref}}(y_w|x)} \right)}{\exp \left( \beta \log \frac{\pi\theta(y_{w}|x)}{\pi_{\text{ref}}(y_{w}|x)} \right)}\\ \\
    + \frac{ \sum_{i=1}^{K-1} \exp \left( \beta \log \frac{\pi_\theta(y_{l_i}|x)}{\pi_{\text{ref}}(y_{l_i}|x)} \right)}{\exp \left( \beta \log \frac{\pi\theta(y_{w}|x)}{\pi_{\text{ref}}(y_{w}|x)} \right)} \right) \right]\\
\label{eqn: base_pl}
\end{dmath}

 Additionally, we also experiment with `Sep-DPO' where each rejected response is treated as a separate data sample paired with the chosen response instead of combining them into one data sample. More information about the formulation of `Sep-DPO' can be found in Appendix ~\ref{sep-dpo}.
\subsection{\textbf{\texttt{TrueBrief.Detection}}}
An advantage of using SLMs is the ability to run them locally and hence access intermediate model outputs. Leveraging this, we extend works by~\cite{logitlens} and~\cite{chuang2024lookbacklensdetectingmitigating} to use the internal representations of \textbf{\texttt{TrueBrief.Finetuned}} models to detect hallucination in their generated summaries. At each token generation step (say $t$), we take the probability of predicting that token across all layers (say $l$), resulting in a feature vector [$LL_{t \times l}$]. We then pool this vector along the token dimension to obtain the \textit{LogitLens vector}, [$LL_{l}$]. This method captures model dynamics during token generation. We also compute the \textit{LookbackLens vector}, which assesses the relative attention attributed to the context at each token generation step. The Lookback ratio is calculated by taking the ratio of average attention weights focused on the context tokens versus the newly generated tokens. The Lookback ratios across all heads (say $h$) and layers (say $l$) are concatenated into a feature vector. This is done at each token generation step (say $t$) to obtain the feature vector, [$LR_{h\times l \times t}$] which is pooled along the token dimension to give us the LookbackLens vector [$LR_{h \times l}$]. We then concatenate the LogitLens and LookbackLens vectors: $[LR_{h\times l}, LL_l]$ and train a simple classifier~\cite{pedregosa2011scikit} for hallucination detection. We find that simple ensembling methods and more complex classifiers are not successful here and concatenation provides a more effective solution. We experiment with LogisticRegression, SVM and NN classifiers from \cite{pedregosa2011scikit}. We also experiment with different pooling strategies for the lens vectors, namely, mean-pooling, max-pooling and statistical-pooling. Table~\ref{tab: detection_grid_results} presents the results from our analysis. We find LogisticRegression with mean-pooling generally gives the best results. 
\begin{figure}[h]
    \centering
    \includegraphics[width=\linewidth]{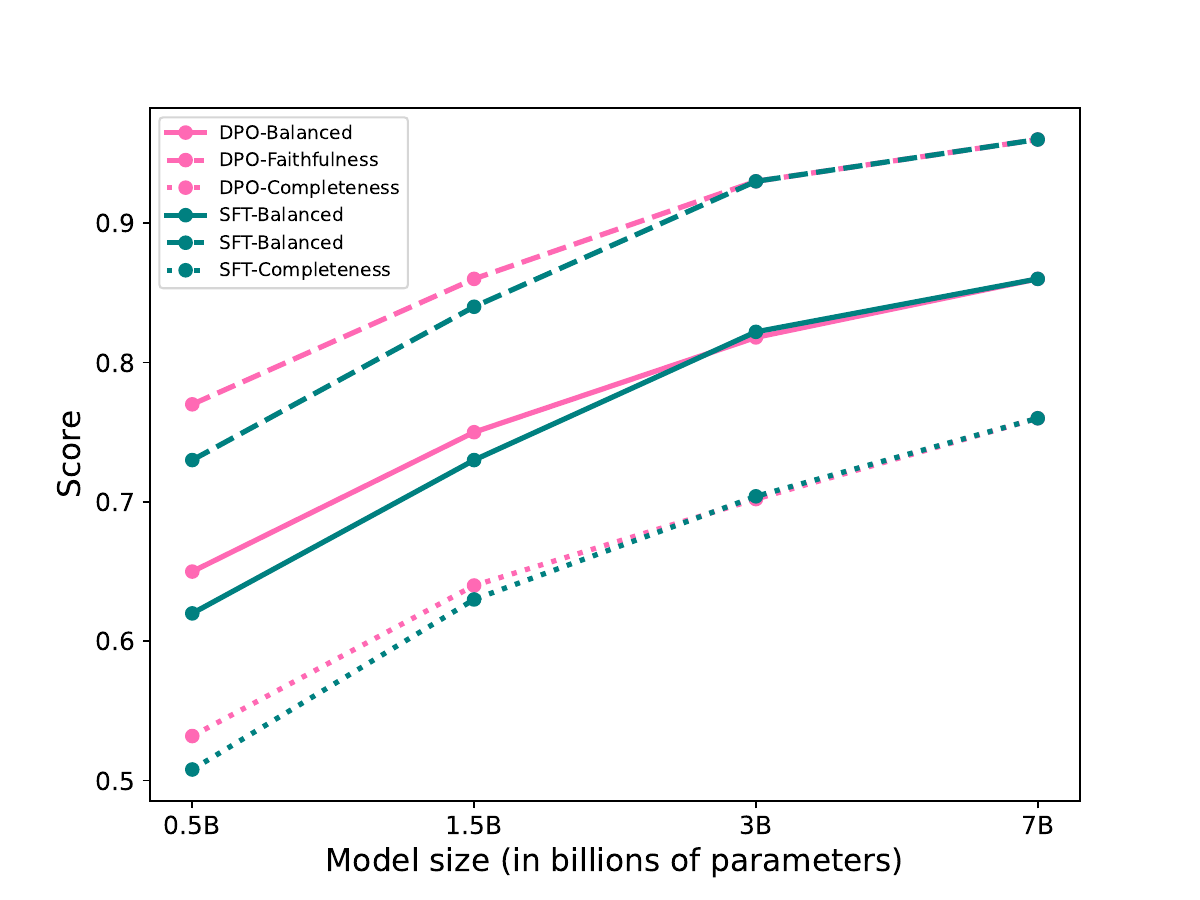}
    \caption{The gap between DPO and SFT reduces as model scale increases which suggests that the effects of preference based optimization are less pronounced in larger models whereas DPO affords significant performance improvement in SLMs}
    \label{fig:scale_vs_performance}
\end{figure} 
\begin{table*}[h]
\centering
\begin{tabular}{lllllllllll}
\hline
 & & \multicolumn{9}{c}{\textbf{Pooling Strategy}} \\
 \hline
\textbf{Model}& \textbf{Classifier} & \multicolumn{3}{c}{\textbf{Mean}} & \multicolumn{3}{c}{\textbf{Max}} & \multicolumn{3}{c}{\textbf{Statistical}} \\
\hline
 & & \textbf{P}&\textbf{R} & \textbf{F1} & \textbf{P}&\textbf{R} & \textbf{F1} & \textbf{P}&\textbf{R} & \textbf{F1}\\
\hline
\multirow{3}{*}{0.5B} & LogisticRegression & 0.31 & 0.75 & 0.44 & 0.27 & 0.60 & 0.37 & 0.30 & 0.68 &0.42\\
 & SVM & 0.27 & 0.82 & 0.41 & 0.25 & 0.78 & 0.38 & 0.26 & 0.80 & 0.40\\
 & NN & 0.30 & 0.72 & 0.42 & 0.28 & 0.72 & 0.40 & 0.27 & 0.82 & 0.41\\
 \hline
 \multirow{3}{*}{1.5B} & LogisticRegression & 0.31 & 0.79 & 0.45 & 0.28 & 0.62 & 0.39 & 0.32 & 0.75 & 0.45\\
  & SVM & 0.29 & 0.86 & 0.43 & 0.27 & 0.80 & 0.41 & 0.29 & 0.86 & 0.44\\
  & NN & 0.29 & 0.79 & 0.43 & 0.29 & 0.70 & 0.41 & 0.32 & 0.78 & 0.46\\
  \hline
  \multirow{3}{*}{3B} & LogisticRegression & 0.32 & 0.77 & 0.45 & 0.29 & 0.57 & 0.38 & 0.33 & 0.74 & 0.45\\
  & SVM& 0.28 & 0.85 & 0.42 & 0.27 & 0.79 & 0.40 & 0.28 & 0.85 & 0.42\\
  & NN & 0.29 & 0.80 & 0.43 & 0.28 & 0.65 & 0.40 & 0.30 & 0.78 & 0.44\\
\hline
\end{tabular}
\caption{\label{tab: detection_grid_results}
TrueBrief.Detection results for grid-search experiments run to select the best classifier and pooling method for different models.
}
\end{table*}
All the classifiers are implemented using the sklearn library~\cite{pedregosa2011scikit}. We set \texttt{max_iter=1000} for all the classifiers. For NN we use hidden layers of sizes 256, 128, 128 and 64 and enable early stopping. We select the best configuration based on F1-Score.
Table~\ref{tab: detection_main_results} shows that using the internal model dynamics from finetuned SLMs is an efficient and low latency alternative for hallucination detection which does not require any API or LLM calls. We are able to achieve reasonable performance using a simple classifier. Although the detection component is designed to work effectively with any model, we observe improved performance when using models finetuned using \textbf{\texttt{TrueBrief.Finetuning}}. We attribute this to the fact that our finetuning module explicitly optimizes for reducing hallucinations which is reflected in the internal model representations of the trained model. Since the detection pipeline relies on these model representations, using a model which favours hallucination-averse behaviour yields the most optimal results. 
\begin{table}[h]
\small
\centering
\begin{tabular}{llll}
\hline
\textbf{Method} & \textbf{P} & \textbf{R} & \textbf{F1} \\
\hline
Prompt_{gpt-3.5-turbo} & 0.23&0.89&0.37\\
Prompt_{gpt-4-turbo} &0.31&\textbf{0.97}&0.47\\
SelfCheckGPT_{gpt-3.5-turbo} & 0.31&0.56&0.40\\
LMvLM_{gpt-4-turbo}&0.23&0.81&0.36\\
Finetuned Llama-2-13b&0.64&0.54&0.59\\
RAG-HAT  (Llama-3-8B-Instruct)& \textbf{0.77} & 0.59 & \textbf{0.67} \\
BERT-base-uncased & 0.23& 0.94& 0.37\\
Longformer-base-4096& 0.27 & 0.65 &  0.38\\
\hdashline
TrueBrief.Detection-0.5B & \textbf{0.34}& 0.45&  0.39\\
TrueBrief.Detection-0.5B (\texttt{TB.F}) & 0.31& 0.75 & 0.44\\
TrueBrief.Detection-1.5B &  0.32&  0.72&  0.44\\
TrueBrief.Detection-1.5B (\texttt{TB.F}) & 0.31 & \textbf{0.79} & \textbf{0.45}\\
TrueBrief.Detection-3B &  0.31&  0.72&  0.43\\
TrueBrief.Detection-3B (\texttt{TB.F}) & 0.32 & 0.77 & \textbf{0.45}\\
\hline
\end{tabular}
\caption{\label{tab: detection_main_results}
Hallucination detection performance on the RAGTruth dataset for the summarization task. \textbf{\texttt{TrueBrief.Detection}} results use the baseline models (without task-specific finetuning) and models finetuned using \textbf{\texttt{TrueBrief.Finetuning}} (\texttt{TB.F}). All \textbf{\texttt{TrueBrief.Detection}} results use the LogisticRegression classifier with the mean pooling strategy for final classification.
}
\end{table}
\section{Experiments}
In this section we provide details pertaining to our experimental set-up including the data, baselines and evaluation metrics.
\subsection{Dataset}
For detection, we use a subset of the RAGTruth dataset containing responses by GPT-4~\cite{openai2024gpt4technicalreport}, GPT-3.5~\cite{openai_gpt35_turbo}, Llama2-70B and Llama2-13B models~\cite{touvron2023llama2openfoundation} (2379 samples) in order to ensure only high quality data is used for training. For the self-detection task, we train the detection model on the same training set, but test using 100 source documents subsampled from the test set. The labels for this test set are defined based on the faithfulness score (see \autoref{subsec:metrics}).
   
\subsection{Implementation Details}
All experiments are carried out on 8 Nvidia-A10g 23 GB GPUs. We use $\beta=0.5$ with an effective batch size of 4 and the AdamW optimizer with a learning rate of 1e-4 by default. For efficient finetuning we utilize LoRA with $\text{rank}=16$, $\text{dropout}=0.05$ and all projection layers as target modules. We use a cosine scheduler with a warmup ratio of 5\%. We train all models for 10 epochs and select the best checkpoint based on the faithfulness score calculated on the validation set. 
\subsection{Metrics}
\label{subsec:metrics}
The duality of our framework warrants separate metrics for assessing the finetuning component and the detection component. The former requires analysing response quality and faithfulness while the latter is treated as a classification task.
\subsubsection{Response Quality and Faithfulness}
We employ Rouge Score~\cite{lin-2004-rouge}, Meteor Score~\cite{meteor} and Bert Score~\cite{zhang2020bertscoreevaluatingtextgeneration} for evaluating response quality based on generated tokens. However, we find these metrics wanting when used for evaluating responses generated by  modern LLMs. They fail to capture the full spectrum of response quality, ignoring aspects like coherence, completeness and fluency. For this reason, we also include evaluations by GPT-4o (2024-08-06)~\cite{openai2024gpt4technicalreport} in an LLM-as-judge setting. The model is also used to calculate a faithfulness score ($F_{score}$) through a pipeline adapted from~\cite{ragas} where we first extract atomic statements and evaluate statement-level faithfulness respectively. We experimentally validate our use of this metric. We then calculate a balanced score by averaging the completeness score and faithfulness score to obtain $B_{score}$ which we use as the primary performance metrics. For the LLM-as-judge metric calculation, the model is asked to score a given summary on a scale of 1 to 5 along the dimensions of coherence, completeness, fluency and relevance. Importantly, this scoring is done with reference to the ground truth summary. Additional implementation details related to the LLM-as-judge setup can be found in ~\ref{app:llm_as_judge}. Futhermore, we use GPT-4o~\cite{openai2024gpt4technicalreport} to score each response for faithfulness through an adapted pipeline similar to ~\cite{ragas} and calculate faithfulness score as:
\begin{dmath}
    F_{score}=\frac{\text{\#faithful_statements}}{\text{\#total_statements}}
\end{dmath}
We find faithfulness to be negatively correlated with response length. Very short summaries would have a high faithfulness score but may not very helpful simply due to the lack of enough information. On the other hand, longer summaries though helpful have more chances of hallucination. Therefore, we use a balanced score which takes into consideration response completeness along with the faithfulness score to ensure the best performing model generates complete and faithful responses. We define the balanced score as follows:
\begin{dmath}
    B_{score}=\frac{C_{score}+F_{score}}{2}
\end{dmath}
where $C_{score}$ is the normalized completeness score from the LLM-as-judge metric and $F_{score}$ is the faithfulness score. We use $B_{score}$ as the final metric for model performance analysis and comparison. Details related to the experiments for validating the reliability of the LLM-based metrics can be found in~\ref{app:metrc_details}

\subsubsection{Hallucination Detection}
We treat hallucination detection as a binary classification task. Following RAGTruth~\cite{niu-etal-2024-ragtruth}, we use the precision, recall and F1-score of the hallucination class to assess performance of the detection models.

\subsection{Models and Baselines}
We use the Qwen2.5 family of models~\cite{qwen2.5} as baselines for the generation task based on their strong performance on public benchmarks as reported here~\cite{open_llm_leaderboard}. To rigorously investigate the impact of model scale while minimizing confounding variables, we use models from the same family (Qwen2.5). With our current set up we can assume the only variation is in model size and keep other variables constant achieving the most stable experimental setting possible in the current landscape. 
Given the scope of this work, we confine our experiments to models with up-to 3 billion parameters, with the exception of a single experiment involving a 7B model. 

For the detection task, we use encoder-based binary classifiers as baselines. Specifically, we report results using BERT~\cite{devlin2019bertpretrainingdeepbidirectional} and Longformer~\cite{beltagy2020longformerlongdocumenttransformer} models. We also use results reported by ~\cite{niu-etal-2024-ragtruth} for additional baselines including LLM prompting, SelfCheckGPT~\cite{manakul2023selfcheckgptzeroresourceblackboxhallucination}, LMvLM~\cite{cohen2023lmvslmdetecting} and LLM finetuning. 

\section{Results and Analysis}

In this section we present the keys insights drawn from our experiments focusing on the effectiveness of \textbf{\texttt{TrueBrief}} in enhancing model faithfulness.

\subsection*{DPO-based faithfulness finetuning is sensitive to data quality}
We observe that the effectiveness of DPO is highly contingent on the quality of the preference data used for finetuning. Table~\ref{tab: data_and_hallucination} shows that using high quality data containing instances of both intrinsic and extrinsic hallucination achieves the highest faithfulness score. However, performance is sensitive to data quality and decreases when significantly higher levels of hallucination are injected into the ground truth data. This sensitivity to data quality further underscores the importance of robust data generation processes, such as the one facilitated by \textbf{\texttt{TrueBrief.DataGen}} to ensure that the preference data effectively represents the desired behaviour.

\subsection*{Less is more in preference-based optimization}
Our findings indicate that in preference-based optimization, using a single rejected response for each chosen response yields the most effective results. We observe this trend across all models as evidenced in table~\ref{tab: finetuning_results}. DPO always out-performs other preference based objectives that use multiple rejected responses such as Add-DPO and PL-DPO. As an ablation, we also experimented with an alternative DPO setting `Sep-DPO' for the 0.5B model (details in Appendix~\ref{sep-dpo}). However, this approach did not outperform DPO. This suggests that additional complexity does not translate into better outcomes. The presence of multiple rejected responses, injects noise and ambiguity, potentially diluting the model's focus on key distinctions between faithful and hallucinated responses. We find that using a single rejected response allows the model to effectively internalize the preference signal, leading to more optimal alignment results. 
Additional analysis related to the effect of model size on effectiveness of DPO is in Appendix~\ref{additional_insights}.

\subsection*{White-box detection using SLMs is an effective and low latency alternative}
\begin{table}[h]
\centering
\begin{tabular}{llll}
\hline
\textbf{Model} & \textbf{Precision} & \textbf{Recall} & \textbf{F1} \\
\hline
TrueBrief-0.5B & \textbf{0.93}& \textbf{0.72}&\textbf{0.81} \\
TrueBrief-1.5B & 0.60& 0.79& 0.68\\
TrueBrief-3.0B & 0.48& 0.68& 0.56\\
\hline
\end{tabular}
\caption{\label{tab: self_detection_results}\textbf{\texttt{TrueBrief.Detection}} performance for hallucination self-detection. All models finetuned using \textbf{\texttt{TrueBrief.Finetuning}}. Test set labels are defined by classifying samples as hallucinated if $F_{score} < 0.9$. Test samples are the same but the distribution of (true, hallucinated) samples varies per model, with (24,76), (48,52) and (69,31) samples for the 0.5B, 1.5B and 3B models respectively.}

\end{table}
Table~\ref{tab: self_detection_results} shows that models trained using \textbf{\texttt{TrueBrief.Finetuning}} are highly proficient at detecting hallucination in their own responses. This self-detection ability declines as model size increases which suggests that hallucinations from more faithful models are more difficult to detect, \textit{even by the models that generated the responses}.

\section{Conclusion}
We propose \textbf{\texttt{TrueBrief}} to enhance the faithfulness of SLMs for summarization. We introduce a data generation method that incorporates controlled hallucination injection to generate high-quality preference-based data. Our experiments, demonstrate that "less is more" showing that a single rejected response is more effective than multiple ones for the summarization task when using preference based optimization. We also provide insights into the optimal use of DPO based on model scale within this experimental setup. Our hallucination detection method uses internal model dynamics, offering a real-time solution without external dependencies. This work underscores the potential of SLMs as effective alternatives to larger models, particularly in security-critical domains, advocating for their broader adoption where resource efficiency is crucial, specifically within the scope of summarization using the Qwen-2.5 models.

\section{Limitations}
The framework's effectiveness on very large models remains unexplored, and the observed diminishing returns with increasing model size suggest that adaptations may be necessary for larger models. As the size of the model increases, finetuning becomes increasingly difficult and computationally expensive therefore limiting the applicability of the proposed method to significantly large models. Additionally, the complexity of controlled hallucination injection in data generation could pose challenges in maintaining consistency and quality across diverse datasets. Moreover, our work primarily provides an experimental analysis, and a comprehensive theoretical exploration of the underlying mechanisms and their implications is still lacking, which could further validate and enhance the framework's applicability. Due to limited computational resources we also did not use larger models for generating the synthetic dataset, this limits the quality of our generated dataset and hence the overall performance of the proposed system.
\bibliography{custom}

@misc{si2023promptinggpt3reliable,
      title={Prompting GPT-3 To Be Reliable}, 
      author={Chenglei Si and Zhe Gan and Zhengyuan Yang and Shuohang Wang and Jianfeng Wang and Jordan Boyd-Graber and Lijuan Wang},
      year={2023},
      eprint={2210.09150},
      archivePrefix={arXiv},
      primaryClass={cs.CL},
      url={https://arxiv.org/abs/2210.09150}, 
}

@misc{varshney2023stitchtimesavesnine,
      title={A Stitch in Time Saves Nine: Detecting and Mitigating Hallucinations of LLMs by Validating Low-Confidence Generation}, 
      author={Neeraj Varshney and Wenlin Yao and Hongming Zhang and Jianshu Chen and Dong Yu},
      year={2023},
      eprint={2307.03987},
      archivePrefix={arXiv},
      primaryClass={cs.CL},
      url={https://arxiv.org/abs/2307.03987}, 
}

@misc{peng2023checkfactstryagain,
      title={Check Your Facts and Try Again: Improving Large Language Models with External Knowledge and Automated Feedback}, 
      author={Baolin Peng and Michel Galley and Pengcheng He and Hao Cheng and Yujia Xie and Yu Hu and Qiuyuan Huang and Lars Liden and Zhou Yu and Weizhu Chen and Jianfeng Gao},
      year={2023},
      eprint={2302.12813},
      archivePrefix={arXiv},
      primaryClass={cs.CL},
      url={https://arxiv.org/abs/2302.12813}, 
}

@inproceedings{gao-etal-2023-rarr,
    title = "{RARR}: Researching and Revising What Language Models Say, Using Language Models",
    author = "Gao, Luyu  and
      Dai, Zhuyun  and
      Pasupat, Panupong  and
      Chen, Anthony  and
      Chaganty, Arun Tejasvi  and
      Fan, Yicheng  and
      Zhao, Vincent  and
      Lao, Ni  and
      Lee, Hongrae  and
      Juan, Da-Cheng  and
      Guu, Kelvin",
    editor = "Rogers, Anna  and
      Boyd-Graber, Jordan  and
      Okazaki, Naoaki",
    booktitle = "Proceedings of the 61st Annual Meeting of the Association for Computational Linguistics (Volume 1: Long Papers)",
    month = jul,
    year = "2023",
    address = "Toronto, Canada",
    publisher = "Association for Computational Linguistics",
    url = "https://aclanthology.org/2023.acl-long.910",
    doi = "10.18653/v1/2023.acl-long.910",
    pages = "16477--16508",
}

@inproceedings{semnani-etal-2023-wikichat,
    title = "{W}iki{C}hat: Stopping the Hallucination of Large Language Model Chatbots by Few-Shot Grounding on {W}ikipedia",
    author = "Semnani, Sina  and
      Yao, Violet  and
      Zhang, Heidi  and
      Lam, Monica",
    editor = "Bouamor, Houda  and
      Pino, Juan  and
      Bali, Kalika",
    booktitle = "Findings of the Association for Computational Linguistics: EMNLP 2023",
    month = dec,
    year = "2023",
    address = "Singapore",
    publisher = "Association for Computational Linguistics",
    url = "https://aclanthology.org/2023.findings-emnlp.157",
    doi = "10.18653/v1/2023.findings-emnlp.157",
    pages = "2387--2413",
}

@inproceedings{shuster-etal-2021-retrieval-augmentation,
    title = "Retrieval Augmentation Reduces Hallucination in Conversation",
    author = "Shuster, Kurt  and
      Poff, Spencer  and
      Chen, Moya  and
      Kiela, Douwe  and
      Weston, Jason",
    editor = "Moens, Marie-Francine  and
      Huang, Xuanjing  and
      Specia, Lucia  and
      Yih, Scott Wen-tau",
    booktitle = "Findings of the Association for Computational Linguistics: EMNLP 2021",
    month = nov,
    year = "2021",
    address = "Punta Cana, Dominican Republic",
    publisher = "Association for Computational Linguistics",
    url = "https://aclanthology.org/2021.findings-emnlp.320",
    doi = "10.18653/v1/2021.findings-emnlp.320",
    pages = "3784--3803",
}

@misc{manakul2023selfcheckgptzeroresourceblackboxhallucination,
      title={SelfCheckGPT: Zero-Resource Black-Box Hallucination Detection for Generative Large Language Models}, 
      author={Potsawee Manakul and Adian Liusie and Mark J. F. Gales},
      year={2023},
      eprint={2303.08896},
      archivePrefix={arXiv},
      primaryClass={cs.CL},
      url={https://arxiv.org/abs/2303.08896}, 
}

@inproceedings{dhuliawala-etal-2024-chain,
    title = "Chain-of-Verification Reduces Hallucination in Large Language Models",
    author = "Dhuliawala, Shehzaad  and
      Komeili, Mojtaba  and
      Xu, Jing  and
      Raileanu, Roberta  and
      Li, Xian  and
      Celikyilmaz, Asli  and
      Weston, Jason",
    editor = "Ku, Lun-Wei  and
      Martins, Andre  and
      Srikumar, Vivek",
    booktitle = "Findings of the Association for Computational Linguistics: ACL 2024",
    month = aug,
    year = "2024",
    address = "Bangkok, Thailand",
    publisher = "Association for Computational Linguistics",
    url = "https://aclanthology.org/2024.findings-acl.212",
    doi = "10.18653/v1/2024.findings-acl.212",
    pages = "3563--3578",
}

@inproceedings{ji-etal-2023-towards,
    title = "Towards Mitigating {LLM} Hallucination via Self Reflection",
    author = "Ji, Ziwei  and
      Yu, Tiezheng  and
      Xu, Yan  and
      Lee, Nayeon  and
      Ishii, Etsuko  and
      Fung, Pascale",
    editor = "Bouamor, Houda  and
      Pino, Juan  and
      Bali, Kalika",
    booktitle = "Findings of the Association for Computational Linguistics: EMNLP 2023",
    month = dec,
    year = "2023",
    address = "Singapore",
    publisher = "Association for Computational Linguistics",
    url = "https://aclanthology.org/2023.findings-emnlp.123",
    doi = "10.18653/v1/2023.findings-emnlp.123",
    pages = "1827--1843",
}

@misc{jones2023teachinglanguagemodelshallucinate,
      title={Teaching Language Models to Hallucinate Less with Synthetic Tasks}, 
      author={Erik Jones and Hamid Palangi and Clarisse Simões and Varun Chandrasekaran and Subhabrata Mukherjee and Arindam Mitra and Ahmed Awadallah and Ece Kamar},
      year={2023},
      eprint={2310.06827},
      archivePrefix={arXiv},
      primaryClass={cs.CL},
      url={https://arxiv.org/abs/2310.06827}, 
}

@inproceedings{cheng-etal-2023-uprise,
    title = "{UPRISE}: Universal Prompt Retrieval for Improving Zero-Shot Evaluation",
    author = "Cheng, Daixuan  and
      Huang, Shaohan  and
      Bi, Junyu  and
      Zhan, Yuefeng  and
      Liu, Jianfeng  and
      Wang, Yujing  and
      Sun, Hao  and
      Wei, Furu  and
      Deng, Weiwei  and
      Zhang, Qi",
    editor = "Bouamor, Houda  and
      Pino, Juan  and
      Bali, Kalika",
    booktitle = "Proceedings of the 2023 Conference on Empirical Methods in Natural Language Processing",
    month = dec,
    year = "2023",
    address = "Singapore",
    publisher = "Association for Computational Linguistics",
    url = "https://aclanthology.org/2023.emnlp-main.758",
    doi = "10.18653/v1/2023.emnlp-main.758",
    pages = "12318--12337",
}

@misc{hu2024mitigatinglargelanguagemodel,
      title={Mitigating Large Language Model Hallucination with Faithful Finetuning}, 
      author={Minda Hu and Bowei He and Yufei Wang and Liangyou Li and Chen Ma and Irwin King},
      year={2024},
      eprint={2406.11267},
      archivePrefix={arXiv},
      primaryClass={cs.CL},
      url={https://arxiv.org/abs/2406.11267}, 
}

@misc{tian2023finetuninglanguagemodelsfactuality,
      title={Fine-tuning Language Models for Factuality}, 
      author={Katherine Tian and Eric Mitchell and Huaxiu Yao and Christopher D. Manning and Chelsea Finn},
      year={2023},
      eprint={2311.08401},
      archivePrefix={arXiv},
      primaryClass={cs.CL},
      url={https://arxiv.org/abs/2311.08401}, 
}

@misc{yang2024alignmenthonesty,
      title={Alignment for Honesty}, 
      author={Yuqing Yang and Ethan Chern and Xipeng Qiu and Graham Neubig and Pengfei Liu},
      year={2024},
      eprint={2312.07000},
      archivePrefix={arXiv},
      primaryClass={cs.CL},
      url={https://arxiv.org/abs/2312.07000}, 
}

@inproceedings{zhang-etal-2024-r,
    title = "{R}-Tuning: Instructing Large Language Models to Say {`}{I} Don{'}t Know{'}",
    author = {Zhang, Hanning, et al.},
    editor = "Duh, Kevin  and
      Gomez, Helena  and
      Bethard, Steven",
    booktitle = "Proceedings of the 2024 Conference of the North American Chapter of the Association for Computational Linguistics: Human Language Technologies (Volume 1: Long Papers)",
    month = jun,
    year = "2024",
    address = "Mexico City, Mexico",
    publisher = "Association for Computational Linguistics",
    url = "https://aclanthology.org/2024.naacl-long.394",
    doi = "10.18653/v1/2024.naacl-long.394",
    pages = "7113--7139",
}

@misc{sun2023aligninglargemultimodalmodels,
      title={Aligning Large Multimodal Models with Factually Augmented RLHF}, 
      author={Zhiqing Sun and Sheng Shen and Shengcao Cao and Haotian Liu and Chunyuan Li and Yikang Shen and Chuang Gan and Liang-Yan Gui and Yu-Xiong Wang and Yiming Yang and Kurt Keutzer and Trevor Darrell},
      year={2023},
      eprint={2309.14525},
      archivePrefix={arXiv},
      primaryClass={cs.CV},
      url={https://arxiv.org/abs/2309.14525}, 
}

@misc{mesgar2021improvingfactualconsistencyresponse,
      title={Improving Factual Consistency Between a Response and Persona Facts}, 
      author={Mohsen Mesgar and Edwin Simpson and Iryna Gurevych},
      year={2021},
      eprint={2005.00036},
      archivePrefix={arXiv},
      primaryClass={cs.CL},
      url={https://arxiv.org/abs/2005.00036}, 
}

@inproceedings{roit-etal-2023-factually,
    title = "Factually Consistent Summarization via Reinforcement Learning with Textual Entailment Feedback",
    author = {Roit, Paul, et al.} ,
    editor = "Rogers, Anna  and
      Boyd-Graber, Jordan  and
      Okazaki, Naoaki",
    booktitle = "Proceedings of the 61st Annual Meeting of the Association for Computational Linguistics (Volume 1: Long Papers)",
    month = jul,
    year = "2023",
    address = "Toronto, Canada",
    publisher = "Association for Computational Linguistics",
    url = "https://aclanthology.org/2023.acl-long.344",
    doi = "10.18653/v1/2023.acl-long.344",
    pages = "6252--6272",
}

@misc{kang2024unfamiliarfinetuningexamplescontrol,
      title={Unfamiliar Finetuning Examples Control How Language Models Hallucinate}, 
      author={Katie Kang and Eric Wallace and Claire Tomlin and Aviral Kumar and Sergey Levine},
      year={2024},
      eprint={2403.05612},
      archivePrefix={arXiv},
      primaryClass={cs.LG},
      url={https://arxiv.org/abs/2403.05612}, 
}

@misc{rafailov2024directpreferenceoptimizationlanguage,
      title={Direct Preference Optimization: Your Language Model is Secretly a Reward Model}, 
      author={Rafael Rafailov and Archit Sharma and Eric Mitchell and Stefano Ermon and Christopher D. Manning and Chelsea Finn},
      year={2024},
      eprint={2305.18290},
      archivePrefix={arXiv},
      primaryClass={cs.LG},
      url={https://arxiv.org/abs/2305.18290}, 
}

@book{luce1959individual,
  title={Individual choice behavior},
  author={Luce, R Duncan},
  volume={4},
  year={1959},
  publisher={Wiley New York}
}

@article{plackett,
 ISSN = {00359254, 14679876},
 URL = {http://www.jstor.org/stable/2346567},
 author = {R. L. Plackett},
 journal = {Journal of the Royal Statistical Society. Series C (Applied Statistics)},
 number = {2},
 pages = {193--202},
 publisher = {[Royal Statistical Society, Oxford University Press]},
 title = {The Analysis of Permutations},
 urldate = {2024-12-04},
 volume = {24},
 year = {1975}
}

@article{bradley-terry,
 ISSN = {00063444, 14643510},
 URL = {http://www.jstor.org/stable/2334029},
 author = {Ralph Allan Bradley and Milton E. Terry},
 journal = {Biometrika},
 number = {3/4},
 pages = {324--345},
 publisher = {[Oxford University Press, Biometrika Trust]},
 title = {Rank Analysis of Incomplete Block Designs: I. The Method of Paired Comparisons},
 urldate = {2024-12-06},
 volume = {39},
 year = {1952}
}

@inproceedings{shi-etal-2024-trusting,
    title = "Trusting Your Evidence: Hallucinate Less with Context-aware Decoding",
    author = "Shi, Weijia  and
      Han, Xiaochuang  and
      Lewis, Mike  and
      Tsvetkov, Yulia  and
      Zettlemoyer, Luke  and
      Yih, Wen-tau",
    editor = "Duh, Kevin  and
      Gomez, Helena  and
      Bethard, Steven",
    booktitle = "Proceedings of the 2024 Conference of the North American Chapter of the Association for Computational Linguistics: Human Language Technologies (Volume 2: Short Papers)",
    month = jun,
    year = "2024",
    address = "Mexico City, Mexico",
    publisher = "Association for Computational Linguistics",
    url = "https://aclanthology.org/2024.naacl-short.69",
    doi = "10.18653/v1/2024.naacl-short.69",
    pages = "783--791",
}

@misc{kossen2024semanticentropyprobesrobust,
      title={Semantic Entropy Probes: Robust and Cheap Hallucination Detection in LLMs}, 
      author={Jannik Kossen and Jiatong Han and Muhammed Razzak and Lisa Schut and Shreshth Malik and Yarin Gal},
      year={2024},
      eprint={2406.15927},
      archivePrefix={arXiv},
      primaryClass={cs.CL},
      url={https://arxiv.org/abs/2406.15927}, 
}

@inproceedings{ch-wang-etal-2024-androids,
    title = "Do Androids Know They{'}re Only Dreaming of Electric Sheep?",
    author = "CH-Wang, Sky  and
      Van Durme, Benjamin  and
      Eisner, Jason  and
      Kedzie, Chris",
    editor = "Ku, Lun-Wei  and
      Martins, Andre  and
      Srikumar, Vivek",
    booktitle = "Findings of the Association for Computational Linguistics: ACL 2024",
    month = aug,
    year = "2024",
    address = "Bangkok, Thailand",
    publisher = "Association for Computational Linguistics",
    url = "https://aclanthology.org/2024.findings-acl.260",
    doi = "10.18653/v1/2024.findings-acl.260",
    pages = "4401--4420",
}

@inproceedings{niu-etal-2024-ragtruth,
    title = "{RAGT}ruth: A Hallucination Corpus for Developing Trustworthy Retrieval-Augmented Language Models",
    author = "Niu, Cheng  and
      Wu, Yuanhao  and
      Zhu, Juno  and
      Xu, Siliang  and
      Shum, KaShun  and
      Zhong, Randy  and
      Song, Juntong  and
      Zhang, Tong",
    editor = "Ku, Lun-Wei  and
      Martins, Andre  and
      Srikumar, Vivek",
    booktitle = "Proceedings of the 62nd Annual Meeting of the Association for Computational Linguistics (Volume 1: Long Papers)",
    month = aug,
    year = "2024",
    address = "Bangkok, Thailand",
    publisher = "Association for Computational Linguistics",
    url = "https://aclanthology.org/2024.acl-long.585",
    doi = "10.18653/v1/2024.acl-long.585",
    pages = "10862--10878",
}

@inproceedings{lin-2004-rouge,
    title = "{ROUGE}: A Package for Automatic Evaluation of Summaries",
    author = "Lin, Chin-Yew",
    booktitle = "Text Summarization Branches Out",
    month = jul,
    year = "2004",
    address = "Barcelona, Spain",
    publisher = "Association for Computational Linguistics",
    url = "https://aclanthology.org/W04-1013",
    pages = "74--81",
}

@inproceedings{meteor,
author = {Lavie, Alon and Agarwal, Abhaya},
title = {Meteor: an automatic metric for MT evaluation with high levels of correlation with human judgments},
year = {2007},
publisher = {Association for Computational Linguistics},
address = {USA},
booktitle = {Proceedings of the Second Workshop on Statistical Machine Translation},
pages = {228–231},
numpages = {4},
location = {Prague, Czech Republic},
series = {StatMT '07}
}

@misc{zhang2020bertscoreevaluatingtextgeneration,
      title={BERTScore: Evaluating Text Generation with BERT}, 
      author={Tianyi Zhang and Varsha Kishore and Felix Wu and Kilian Q. Weinberger and Yoav Artzi},
      year={2020},
      eprint={1904.09675},
      archivePrefix={arXiv},
      primaryClass={cs.CL},
      url={https://arxiv.org/abs/1904.09675}, 
}

@misc{wu2024betadpodirectpreferenceoptimization,
      title={$\beta$-DPO: Direct Preference Optimization with Dynamic $\beta$}, 
      author={Junkang Wu and Yuexiang Xie and Zhengyi Yang and Jiancan Wu and Jinyang Gao and Bolin Ding and Xiang Wang and Xiangnan He},
      year={2024},
      eprint={2407.08639},
      archivePrefix={arXiv},
      primaryClass={cs.AI},
      url={https://arxiv.org/abs/2407.08639}, 
}

@misc{abdin2024phi3technicalreporthighly,
      title={Phi-3 Technical Report: A Highly Capable Language Model Locally on Your Phone}, 
      author={Marah Abdin and team},
      year={2024},
      eprint={2404.14219},
      archivePrefix={arXiv},
      primaryClass={cs.CL},
      url={https://arxiv.org/abs/2404.14219}, 
}

@misc{open_llm_leaderboard,
  title = {Open LLM Leaderboard},
  author={HuggingFace},
  howpublished = {\url{https://huggingface.co/spaces/open-llm-leaderboard/open_llm_leaderboard}},
  note = {Accessed: 2024-12-18},
  year = {2024}
}

@misc{wang2024comprehensivesurveysmalllanguage,
      title={A Comprehensive Survey of Small Language Models in the Era of Large Language Models: Techniques, Enhancements, Applications, Collaboration with LLMs, and Trustworthiness}, 
      author={Fali Wang and Zhiwei Zhang and Xianren Zhang and Zongyu Wu and Tzuhao Mo and Qiuhao Lu and Wanjing Wang and Rui Li and Junjie Xu and Xianfeng Tang and Qi He and Yao Ma and Ming Huang and Suhang Wang},
      year={2024},
      eprint={2411.03350},
      archivePrefix={arXiv},
      primaryClass={cs.CL},
      url={https://arxiv.org/abs/2411.03350}, 
}

@inproceedings{projected-language-models,
title = {Projected Language Models: A Large Model Pre-Segmented Into Smaller Ones},
booktitle = {ICML Workshop},
author = {David Grangier and Angelos Katharopoulos and Pierre Ablin and Awni Hannun},
year = {2024},
URL = {https://openreview.net/forum?id=Wi88giKi7N}
}

@misc{xu2024hallucinationinevitableinnatelimitation,
      title={Hallucination is Inevitable: An Innate Limitation of Large Language Models}, 
      author={Ziwei Xu and Sanjay Jain and Mohan Kankanhalli},
      year={2024},
      eprint={2401.11817},
      archivePrefix={arXiv},
      primaryClass={cs.CL},
      url={https://arxiv.org/abs/2401.11817}, 
}

@article{llm_hallucination_survey,
author = {Huang, Lei and Yu, Weijiang and Ma, Weitao and Zhong, Weihong and Feng, Zhangyin and Wang, Haotian and Chen, Qianglong and Peng, Weihua and Feng, Xiaocheng and Qin, Bing and Liu, Ting},
title = {A Survey on Hallucination in Large Language Models: Principles, Taxonomy, Challenges, and Open Questions},
year = {2024},
publisher = {Association for Computing Machinery},
address = {New York, NY, USA},
issn = {1046-8188},
url = {https://doi.org/10.1145/3703155},
doi = {10.1145/3703155},
note = {Just Accepted},
journal = {ACM Trans. Inf. Syst.},
month = nov
}

@inproceedings{calibrated_hallucination,
author = {Kalai, Adam Tauman and Vempala, Santosh S.},
title = {Calibrated Language Models Must Hallucinate},
year = {2024},
isbn = {9798400703836},
publisher = {Association for Computing Machinery},
address = {New York, NY, USA},
url = {https://doi.org/10.1145/3618260.3649777},
doi = {10.1145/3618260.3649777},
booktitle = {Proceedings of the 56th Annual ACM Symposium on Theory of Computing},
pages = {160–171},
numpages = {12},
location = {Vancouver, BC, Canada},
series = {STOC 2024}
}

@misc{touvron2023llama2openfoundation,
      title={Llama 2: Open Foundation and Fine-Tuned Chat Models}, 
      author={Hugo Touvron and team},
      year={2023},
      eprint={2307.09288},
      archivePrefix={arXiv},
      primaryClass={cs.CL},
      url={https://arxiv.org/abs/2307.09288}, 
}

@misc{openai_gpt35_turbo,
  author = {OpenAI},
  title = {GPT-3.5 Turbo Fine-tuning and API Updates},
  howpublished = {\url{https://openai.com/index/gpt-3-5-turbo-fine-tuning-and-api-updates/}},
  note = {Accessed: 2023-10-21},
  year = {2023}
}

@INPROCEEDINGS{slms_cheaper,
  author={Irugalbandara, Chandra and Mahendra, Ashish and Daynauth, Roland and Arachchige, Tharuka Kasthuri and Dantanarayana, Jayanaka and Flautner, Krisztian and Tang, Lingjia and Kang, Yiping and Mars, Jason},
  booktitle={2024 IEEE International Symposium on Performance Analysis of Systems and Software (ISPASS)}, 
  title={Scaling Down to Scale Up: A Cost-Benefit Analysis of Replacing OpenAI's LLM with Open Source SLMs in Production}, 
  year={2024},
  volume={},
  number={},
  pages={280-291},
  doi={10.1109/ISPASS61541.2024.00034}}

@misc{chuang2024lookbacklensdetectingmitigating,
      title={Lookback Lens: Detecting and Mitigating Contextual Hallucinations in Large Language Models Using Only Attention Maps}, 
      author={Yung-Sung Chuang and Linlu Qiu and Cheng-Yu Hsieh and Ranjay Krishna and Yoon Kim and James Glass},
      year={2024},
      eprint={2407.07071},
      archivePrefix={arXiv},
      primaryClass={cs.CL},
      url={https://arxiv.org/abs/2407.07071}, 
}

@misc{grattafiori2024llama3herdmodels,
      title={The Llama 3 Herd of Models}, 
      author={Llama Team, AI@Meta},
      year={2024},
      eprint={2407.21783},
      archivePrefix={arXiv},
      primaryClass={cs.AI},
      url={https://arxiv.org/abs/2407.21783}, 
}

@misc{cheng2024smallagentrockempowering,
      title={Small Agent Can Also Rock! Empowering Small Language Models as Hallucination Detector}, 
      author={Xiaoxue Cheng and Junyi Li and Wayne Xin Zhao and Hongzhi Zhang and Fuzheng Zhang and Di Zhang and Kun Gai and Ji-Rong Wen},
      year={2024},
      eprint={2406.11277},
      archivePrefix={arXiv},
      primaryClass={cs.CL},
      url={https://arxiv.org/abs/2406.11277}, 
}

@misc{chern2023factoolfactualitydetectiongenerative,
      title={FacTool: Factuality Detection in Generative AI -- A Tool Augmented Framework for Multi-Task and Multi-Domain Scenarios}, 
      author={I-Chun Chern and Steffi Chern and Shiqi Chen and Weizhe Yuan and Kehua Feng and Chunting Zhou and Junxian He and Graham Neubig and Pengfei Liu},
      year={2023},
      eprint={2307.13528},
      archivePrefix={arXiv},
      primaryClass={cs.CL},
      url={https://arxiv.org/abs/2307.13528}, 
}

@misc{min2023factscorefinegrainedatomicevaluation,
      title={FActScore: Fine-grained Atomic Evaluation of Factual Precision in Long Form Text Generation}, 
      author={Sewon Min and Kalpesh Krishna and Xinxi Lyu and Mike Lewis and Wen-tau Yih and Pang Wei Koh and Mohit Iyyer and Luke Zettlemoyer and Hannaneh Hajishirzi},
      year={2023},
      eprint={2305.14251},
      archivePrefix={arXiv},
      primaryClass={cs.CL},
      url={https://arxiv.org/abs/2305.14251}, 
}

@article{qwen2.5,
    title   = {Qwen2.5 Technical Report}, 
    author  = {An Yang and team},
    journal = {arXiv preprint arXiv:2412.15115},
    year    = {2024}
}

@misc{wang2024directjudgementpreferenceoptimization,
      title={Direct Judgement Preference Optimization}, 
      author={Peifeng Wang and Austin Xu and Yilun Zhou and Caiming Xiong and Shafiq Joty},
      year={2024},
      eprint={2409.14664},
      archivePrefix={arXiv},
      primaryClass={cs.CL},
      url={https://arxiv.org/abs/2409.14664}, 
}

@inproceedings{raghat,
    title = "{RAG}-{HAT}: A Hallucination-Aware Tuning Pipeline for {LLM} in Retrieval-Augmented Generation",
    author = "Song, Juntong  and
      Wang, Xingguang  and
      Zhu, Juno  and
      Wu, Yuanhao  and
      Cheng, Xuxin  and
      Zhong, Randy  and
      Niu, Cheng",
    editor = "Dernoncourt, Franck  and
      Preo{\c{t}}iuc-Pietro, Daniel  and
      Shimorina, Anastasia",
    booktitle = "Proceedings of the 2024 Conference on Empirical Methods in Natural Language Processing: Industry Track",
    month = nov,
    year = "2024",
    address = "Miami, Florida, US",
    publisher = "Association for Computational Linguistics",
    url = "https://aclanthology.org/2024.emnlp-industry.113/",
    doi = "10.18653/v1/2024.emnlp-industry.113",
    pages = "1548--1558"
}

@misc{belyi2024lunaevaluationfoundationmodel,
      title={Luna: An Evaluation Foundation Model to Catch Language Model Hallucinations with High Accuracy and Low Cost}, 
      author={Masha Belyi and Robert Friel and Shuai Shao and Atindriyo Sanyal},
      year={2024},
      eprint={2406.00975},
      archivePrefix={arXiv},
      primaryClass={cs.CL},
      url={https://arxiv.org/abs/2406.00975}, 
}

@misc{cohen2023lmvslmdetecting,
      title={LM vs LM: Detecting Factual Errors via Cross Examination}, 
      author={Roi Cohen and May Hamri and Mor Geva and Amir Globerson},
      year={2023},
      eprint={2305.13281},
      archivePrefix={arXiv},
      primaryClass={cs.CL},
      url={https://arxiv.org/abs/2305.13281}, 
}

@misc{logitlens, title={Interpreting GPT: The logit lens}, url={https://www.lesswrong.com/posts/AcKRB8wDpdaN6v6ru/interpreting-gpt-the-logit-lens}, journal={LessWrong}, author={Nostalgebraist}}

@misc{devlin2019bertpretrainingdeepbidirectional,
      title={BERT: Pre-training of Deep Bidirectional Transformers for Language Understanding}, 
      author={Jacob Devlin and Ming-Wei Chang and Kenton Lee and Kristina Toutanova},
      year={2019},
      eprint={1810.04805},
      archivePrefix={arXiv},
      primaryClass={cs.CL},
      url={https://arxiv.org/abs/1810.04805}, 
}

@misc{beltagy2020longformerlongdocumenttransformer,
      title={Longformer: The Long-Document Transformer}, 
      author={Iz Beltagy and Matthew E. Peters and Arman Cohan},
      year={2020},
      eprint={2004.05150},
      archivePrefix={arXiv},
      primaryClass={cs.CL},
      url={https://arxiv.org/abs/2004.05150}, 
}

@article{pedregosa2011scikit,
  title={Scikit-learn: Machine learning in Python},
  author={Pedregosa, Fabian and Varoquaux, Ga{\"e}l and Gramfort, Alexandre and Michel, Vincent and Thirion, Bertrand and Grisel, Olivier and Blondel, Mathieu and Prettenhofer, Peter and Weiss, Ron and Dubourg, Vincent and others},
  journal={Journal of machine learning research},
  volume={12},
  number={Oct},
  pages={2825--2830},
  year={2011}
}

@misc{ragas, title={Explodinggradients/Ragas: Supercharge Your LLM Application Evaluations}, url={https://github.com/explodinggradients/ragas}, journal={Ragas}, author={Ragas, Exploding Gradients}}

@misc{gpt3,
      title={Language Models are Few-Shot Learners}, 
      author={Tom B. Brown and Benjamin Mann and Nick Ryder and Melanie Subbiah and Jared Kaplan and Prafulla Dhariwal and Arvind Neelakantan and Pranav Shyam and Girish Sastry and Amanda Askell and Sandhini Agarwal and Ariel Herbert-Voss and Gretchen Krueger and Tom Henighan and Rewon Child and Aditya Ramesh and Daniel M. Ziegler and Jeffrey Wu and Clemens Winter and Christopher Hesse and Mark Chen and Eric Sigler and Mateusz Litwin and Scott Gray and Benjamin Chess and Jack Clark and Christopher Berner and Sam McCandlish and Alec Radford and Ilya Sutskever and Dario Amodei},
      year={2020},
      eprint={2005.14165},
      archivePrefix={arXiv},
      primaryClass={cs.CL},
      url={https://arxiv.org/abs/2005.14165}, 
}

@misc{openai2024gpt4technicalreport,
      title={GPT-4 Technical Report}, 
      author={OpenAI},
      year={2024},
      eprint={2303.08774},
      archivePrefix={arXiv},
      primaryClass={cs.CL},
      url={https://arxiv.org/abs/2303.08774}, 
}

@unpublished{spacy2,
    AUTHOR = {Honnibal, Matthew and Montani, Ines},
    TITLE  = {{spaCy 2}: Natural language understanding with {B}loom embeddings, convolutional neural networks and incremental parsing},
    YEAR   = {2017},
    Note   = {To appear}
}

\appendix
\label{appendix}
\onecolumn
\section{Experimental Validation of Reliability of LLM-based Metrics}
\label{app:metrc_details}
\label{app: f_score}
We run the faithfulness score metric as used in the paper on the RAGTruth summarization test data~\cite{niu-etal-2024-ragtruth}. We observe that the faithfulness score is considerably higher for the hallucination-free samples and lower for those with hallucination. While we could not run the scoring on all samples in the test set due to content filter triggering, we observe that the metric is correlated with human-annotated faithfulness i.e. existence of hallucination.

\begin{table*}[h]
    \centering
    \begin{tabular}{lllllllll}
    \hline
    \textbf{Sample Type} & \textbf{Count} & \multicolumn{7}{c}{\textbf{F_{score}}}\\
    & & Mean & Std & Min & 25\% & 50\% & 75\% & Max \\
    \hline
         Hallucinated &  185 & 0.81 & 0.14 & 0.00 & 0.75 & 0.83 & 0.88 & 1.00\\
         Hallucination-Free & 651 & 0.97 & 0.07 & 0.50 & 1.00 & 1.00 & 1.00 & 1.00\\
         \hline
    \end{tabular}
    \caption{F_{score} on human-annotated RAGTruth summarization test set.}
    \label{tab:my_label}
\end{table*}

\section{Complete Table for Finetuning Results}
\label{app:complete_finetuning_results}
This is the complete version of table~\ref{tab: finetuning_results} that includes standard deviation values for the LLM based metrics.
\begin{table*}[h]
\scriptsize
\centering
\begin{tabular}{llllllllllll}
\hline
\textbf{Model} & \textbf{Model} & \multicolumn{3}{c}{\textbf{Rouge Score}} & \textbf{Bert} & \multicolumn{4}{c}{\textbf{LLM as Judge}} & \textbf{Faithful-} &\textbf{Balanced}\\
\textbf{Size}& &\textbf{R-1} & \textbf{R-2} & \textbf{R-L} &\textbf{Score}&\textbf{Completeness} & \textbf{Relevance} & \textbf{Fluency} & \textbf{Coherence}&  \textbf{ness Score} & \textbf{Score}\\
\hline
\multirow{5}{*}{0.5B} & Baseline & 0.40 & 0.16 & 0.25 & 0.64 &2.57$\pm$ 0.14 & 3.18$\pm$0.15  &\textbf{3.92$\pm$0.24}&\textbf{3.60$\pm$ 0.2}2&0.74$\pm$0.01& 0.62 \\
& SFT & 0.44 & \textbf{0.19} & \textbf{0.29} &0.66 &2.54$\pm$0.15&3.15$\pm$0.17&3.91$\pm$0.22&3.52$\pm$0.25& 0.75$\pm$0.00&  0.62 \\
& DPO & 0.44 & \textbf{0.19} & 0.28 & 0.66&\textbf{2.66$\pm$0.19}&\textbf{3.19$\pm$0.22}&3.91$\pm$0.22&3.56$\pm$0.25&\textbf{0.77$\pm$0.04}&\textbf{0.65}\\
& Add-DPO & 0.43 & 0.17 & 0.27&0.65 &2.54$\pm$0.16&3.12$\pm$0.23&3.83$\pm$0.22&3.47$\pm$0.23&0.74$\pm$0.05&0.62\\
& PL-DPO&\textbf{0.45}&0.20&\textbf{0.29}&\textbf{0.67}&2.54$\pm$0.12&3.13$\pm$0.20&3.85$\pm$0.19&3.50$\pm$0.29&0.73$\pm$0.04&0.62\\
\hline
\multirow{5}{*}{1.5B} & Baseline & 0.41 & 0.16 & 0.25 &0.65&3.22$\pm$0.18&3.71$\pm$0.24&\textbf{4.45$\pm$0.24}&\textbf{4.30$\pm$0.27}&0.81$\pm$0.03&0.73\\
& SFT & 0.45 & 0.20 & 0.29 &0.67&3.15$\pm$0.16&3.63$\pm$0.20&4.23$\pm$0.15&4.02$\pm$0.24&0.84$\pm$0.03&0.73\\
& DPO & \textbf{0.46} & 0.20 & \textbf{0.30}&\textbf{0.68}&3.20$\pm$0.18&\textbf{3.81$\pm$0.22}&4.44$\pm$0.16&4.23$\pm$0.20&\textbf{0.86$\pm$0.02}&\textbf{0.75}\\
& Add-DPO & \textbf{0.46} & \textbf{0.21} & \textbf{0.30}&\textbf{0.68} &3.15$\pm$0.15&3.71$\pm$0.19&4.40$\pm$0.19&4.17$\pm$0.18&0.84$\pm$0.02&0.73\\
& PL-DPO & \textbf{0.46} & 0.20 & \textbf{0.30} &\textbf{0.68}&\textbf{3.22$\pm$0.13}&3.67$\pm$0.24&4.37$\pm$0.18&4.16$\pm$0.19&0.85$\pm$0.02&0.74\\
\hline
\multirow{5}{*}{3B}&Baseline & 0.47 & 0.21 & 0.31&0.68 
&3.30$\pm$0.16&\textbf{4.23$\pm$0.19}&4.68$\pm$0.19&4.53$\pm$0.22&0.91$\pm$0.01&0.79\\
& SFT & \textbf{0.52} & \textbf{0.26} & \textbf{0.35} &\textbf{0.71}&\textbf{3.52$\pm$0.16}&4.22$\pm$0.19&4.72$\pm$0.19&\textbf{4.59$\pm$0.18}&\textbf{0.93$\pm$0.01}&\textbf{0.82}\\
& DPO& 0.51& 0.25&0.34&0.70&3.43$\pm$0.16&4.09$\pm$0.22&4.71$\pm$0.13&4.48$\pm$0.19&\textbf{0.93$\pm$0.01}&\textbf{0.82}\\
& Add-DPO&0.51&0.24&0.33&0.70&3.46$\pm$0.17&4.16$\pm$0.24&\textbf{4.73$\pm$0.18}&4.54$\pm$0.24&\textbf{0.93$\pm$0.01}&0.81\\
& PL-DPO&0.51&\textbf{0.26}&0.34&0.70&3.51$\pm$0.17&4.05$\pm$0.24&4.68$\pm$0.16&4.51$\pm$0.15&0.92$\pm$0.01&0.81\\
\hline
\multirow{5}{*}{7B}&Baseline & 0.47 & 0.20 & 0.31 & 0.68&3.30$\pm$0.00&4.24$\pm$0.00&4.73$\pm$0.00&4.60$\pm$0.00&0.95$\pm$0.01&0.80\\
& SFT & \textbf{0.52} & \textbf{0.26} & \textbf{0.35} &\textbf{0.71}& 3.79$\pm$0.00&4.38$\pm$0.00&4.83$\pm$0.00&4.72$\pm$0.00&0.96$\pm$0.01&0.85\\
& DPO & \textbf{0.52}&\textbf{0.26}&\textbf{0.35}&\textbf{0.71}&\textbf{3.80$\pm$0.00}&\textbf{4.39$\pm$0.00}&\textbf{4.85$\pm$0.00}&\textbf{4.73$\pm$0.00}&\textbf{0.96$\pm$0.00}&\textbf{0.86}\\
& Add-DPO &-&-&-&-&-&-&-&-&-&-\\
& PL-DPO &-&-&-&-&-&-&-&-&-&-\\
\hline
\end{tabular}
\caption{\label{tab: full_finetuning_results}
TrueBrief.Finetuning Results: DPO out performs all other methods across models of different sizes. Using multiple rejected responses does not afford any significant gains when using preference based optimization. Moreover, the performance gap between SFT and DPO decreases as model size increases.
}
\end{table*}
\section{DPO excels in smaller models for faithfulness, with diminishing returns in larger models}
\label{additional_insights}
Our analysis reveals that preference-based optimization is effective in enhancing the faithfulness of SLMs (<3B parameters). Table~\ref{tab: finetuning_results} shows that DPO significantly improves the alignment of smaller models, leading to a marked increase in response faithfulness. However, as model size increases, the relative advantage of DPO over standard finetuning techniques diminishes. Figure~\ref{fig:scale_vs_performance} shows that 
DPO is more beneficial for aligning smaller models

\section{System Component Details}
\label{app:sys_components}
\subsection{\textbf{\texttt{TrueBrief.DataGen}}}
The data generator component is used to systematically inject hallucination into the ground truth data to create preference pairs for model finetuning. Intrinsic hallucinations are introduced by factually augmenting entities (extracted through Spacy NER) within the ground truth summary. Extrinsic hallucinations, on the other hand, are introduced by adding external context not derived from the provided source information. This is achieved by prompting Llama-3.1-8B~\cite{grattafiori2024llama3herdmodels} to paraphrase a given sentence, subtly incorporating baseless information while preserving its original meaning. The date generation module contains \texttt{TrueBrief.Hallucinator} and \texttt{TrueBrief.Summarizer} as its constituents. The data generation process employed for the experiments presented in this work involves first factually augmenting the ground truth summary to inject factual hallucination into the data. We extract named entities using SpaCy~\cite{spacy2} and then replace the extracted entities using an LLM with false values. This is followed by systematic paraphrasing of parts of the factually augmented response. We randomly select between at least one sentence to 100\% of the input summary to paraphrase. This process ensures the final hallucinated response does not deviate too far away from the ground truth response in terms of lexical and semantic style. We also ensure that hallucination is always injected into the samples.
\subsection{\texttt{TrueBrief.Hallucinator}}
The hallucinator has two modes for hallucination generation and also makes use of the summarizer to generate certain kinds of hallucination. The three hallucination modes are as follows:
\begin{enumerate}
    \item Factual Hallucination: This mode specifically targets the response (summary) instead of the source information. It requires first identifying the named entities in the response and then replacing a proportion of these entities with related but false values. This mode does not make changes to the structure of the ground truth response.
    \item Paraphrasing Hallucination: In this mode we prompt the LLM to paraphrase parts of the input text such that the original context of the input text remains the same while introducing subtle baseless hallucinations. Since we only paraphrase parts of the  ground truth response, we are still able to preserve ground truth response structure. This mode of hallucination inject is primarily use for extrinsic hallucination injection into the the ground truth response that involves the addition of unverifiable information.
\end{enumerate}
While DPO only needs singular rejected responses for training, our proposed variations make use of multiple rejected responses. In order to generate multiple rejected responses for the same source information, we vary the various above specified parameters for different hallucination methods. We use the factual+paraphrasing hallucination methods to generate the dataset used for training the models for the experiments in this work. We first factually augment the input text and then paraphrase parts of it to ensure the rejected response contains both intrinsic and extrinsic hallucinations. We generate multiple rejected responses by varying the proportion of paraphrased sentences to be with 1, 50\% or 100\% of the sentences in the ground truth data. We observe that generating the dataset in this manner gives us the best results.

The specific prompts used for the various hallucination methods are as follows:
\begin{tcolorbox}[colback=gray!5!white, colframe=gray!75!black, title=Standard Hallucination Prompt, width=\linewidth]
Complete the given text by adding 5-10 more sentences to the <location> of the text so that the final text <nli_sentiment>s the general context: <text>
\end{tcolorbox}
\begin{tcolorbox}[colback=gray!5!white, colframe=gray!75!black, title=Factual Hallucination Prompt, width=\linewidth]
For each item in the list, provide a close but different value. Output the answer in JSON format, with the key as the item and the value as the new value. 
        Ensure the output is valid JSON. Only output the JSON and nothing else. Items: <list_of_entities_to_augment>
\end{tcolorbox}
\begin{tcolorbox}[colback=gray!5!white, colframe=gray!75!black, title=Paraphrasing Hallucination Prompt, width=\linewidth]
You are a highly skilled paraphrasing agent with an exceptional ability to rephrase sentences while preserving their original meaning and context. Your task is to take the given sentence and transform it into a new version that is both coherent and contextually accurate. Feel free to enhance the sentence with relevant details or insights that align with the original intent, showcasing your ability to enrich the content subtly. Only output the new sentence. Here is the sentence to rephrase: <sentence>
\end{tcolorbox}

\subsection{\texttt{TrueBrief.Summarizer}}
The summarizer is a crucial component of the \textbf{\texttt{TrueBrief}} system. It is used as a part of the data generation and hallucinator modules. It is also used for final summary generation. The component prompts the given model to summarize the provided source information.
\begin{tcolorbox}[colback=gray!5!white, colframe=gray!75!black, title=Summarization Prompt, width=\linewidth]
Summarize the following text in one sentence, ensuring all key points are included without any personal opinions or interpretations: <text>
\end{tcolorbox}
\section{Sep-DPO}
\label{sep-dpo}
Separated-DPO (Sep-DPO) involves treating each rejected response as a separate data sample paired with the chosen response. It can be understood as an experimental setting rather then a separate loss function formulation since it uses the same loss function as DPO but with more data samples. The dataset size if tripled in our case, since we have 3 rejected samples for each chosen samples and Sep-DPO treats each rejected sample as a separate data point with the corresponding chosen sample.
\section{LLM as Judge}
\label{app:llm_as_judge}
We use GPT-4o for our LLM as judge setup. For our specific use-case the model is provided the source text, generated summary and ground truth summary. The model is then asked to assess the quality of the generated summary along five dimensions while considering the source text and the ground truth summary. The final output from the model is a score for each dimension on a scale of 1 to 5 with 1 being the lowest score and 5 being the highest. We define each dimension and score value in the prompt provided to the model. The prompt used is presented below:
\onecolumn
\begin{tcolorbox}[colback=gray!5!white, colframe=gray!75!black, title=LLM as Judge Prompt, width=\linewidth]
You are an unbiased and professional judge for evaluating the quality of conversation summary.
  
  In this task you will be provided the following:
            
            1. a "text" taken from a news article, and correspondingly 
            
            2. a "golden summary" written by human expert which is considered best quality
            
            3. a "test summary" written by a model
            
  And your job is to evaluate the quality of the "test summary" using the following dimensions and criteria and score the test summary along each dimension on a scale of 1-5 according to the score definition provided with each dimension.
  
  The evaluation dimensions and criteria are as follows: 
  
        - Completeness: 
        
            - Definition: This dimension assesses how well the summary captures all the important points and details from the original conversation.
            
            - Valid scores: 1, 2, 3, 4, 5
            
                - Score definition:
                    1: Very Incomplete, the summary misses most of the key points and details.
                    2: Incomplete, the summary captures some key points but misses several important details.
                    3: Moderately Complete, the summary captures many key points but misses some details.
                    4: Mostly Complete, the summary captures most of the key points and details.
                    5: Complete, the summary captures all key points and details.

        - Relevance: 
        
            - Definition: This dimension assesses how well the summary focuses on the important and relevant points of the text without including unnecessary or irrelevant information.
            
            - Valid scores: 1, 2, 3, 4, 5
            
                - Score definition:
                    1: Irrelevant, the summary includes mostly irrelevant information.
                    2: Somewhat Relevant, the summary includes some relevant information but also contains unnecessary details.
                    3: Moderately Relevant, the summary includes relevant information but with noticeable irrelevant details.
                    4: Mostly Relevant, the summary focuses on the important points with minimal irrelevant information.
                    5: Highly Relevant, the summary focuses on the important points with no irrelevant information.
        
        - Coherence:
        
            - Definition: This dimension assesses how well the sentences in the summary logically flow from one to the next, creating a unified and sensible whole.
            
            - Valid scores: 1, 2, 3, 4, 5
            
                - Score definition:
                    1: Not Coherent, the summary is disjointed and lacks logical flow.
                    2: Somewhat Coherent, the summary has some logical flow but is still confusing in parts.
                    3: Moderately Coherent, the summary generally flows well but has noticeable logical issues.
                    4: Mostly Coherent, the summary flows well with minor logical issues.
                    5: Highly Coherent, the summary flows logically and makes complete sense as a whole.
                    
        - Fluency:
        
            - Definition: This dimension assesses how well the words and sentences in the summary flow naturally and smoothly, without awkward phrasing or grammatical errors.
            
            - Valid scores: 1, 2, 3, 4, 5
            
                - Score definition:
                    1: Not Fluent, the summary is awkward and difficult to read.
                    2: Somewhat Fluent, the summary has some awkward phrasing or errors.
                    3: Moderately Fluent, the summary reads well but has noticeable phrasing issues.
                    4: Mostly Fluent, the summary reads well with minor phrasing issues.
                    5: Highly Fluent, the summary reads smoothly and naturally.
    
  You must answer for all the 4 evaluation dimensions.
\end{tcolorbox}
\section{Beta value analysis}
\label{app:beta_value_analysis}

The $\beta$ parameter controls the deviation of the new policy from the base reference policy. It maintains a balance between the original reference model and the new preferences. Lower $\beta$ values correspond to more aggressive updates and higher values support more controlled updates.~\cite{wu2024betadpodirectpreferenceoptimization} show that the value of $\beta$ can be influenced by data quality. To this end, we run an ablation study to select the best value of $\beta$ for our use case. We train the Qwen2.5-1.5B-Instruct model on a subset of the training data using the standard DPO objective. Since response quality does not change drastically with $\beta$ value, we use faithfulness score as the main metric to make our selection of $\beta$. Based on results presented in table~\ref{tab: beta_value_ablation} we observe best performance with $\beta=0.5$ and use this $\beta$ value for all experiments presented in this work.
\begin{table*}[h]
\centering
\begin{tabular}{lllllll}
\hline
\textbf{$\beta$ value} & \multicolumn{3}{c}{\textbf{Rouge Score}} & \textbf{Meteor} & \textbf{Bert} & \textbf{Faithfulness} \\
& \textbf{Rouge-1} & \textbf{Rouge-2} & \textbf{Rouge-L} & \textbf{Score}& \textbf{Score}& \textbf{Score}\\
\hline
0.2 & 0.3707 & 0.1572 & 0.2604 & 0.2579 & 0.6461 & 0.6361 \\
0.3 & 0.3766 & 0.1626 & 0.2622 & 0.2618 & 0.6489 & 0.6343 \\
0.4 & 0.3770 & 0.1615 & 0.2632 & 0.2624 & 0.6500 & 0.6471 \\
0.5 & 0.3727 & 0.1596 & 0.2615 & 0.2592 & 0.6482 & 0.6511 \\
0.6 & 0.3729 & 0.1588 & 0.2626 & 0.2601 & 0.6475 & 0.6475 \\
0.7 & 0.3777 & 0.1601 & 0.2639 & 0.2660 & 0.6494 & 0.6362 \\
0.8 & 0.3750 & 0.1616 & 0.2624 & 0.2608 & 0.6482 & 0.6511 \\
\hline
\end{tabular}

\caption{\label{tab: beta_value_ablation}
$\beta=0.5$ gives the best performance in terms of faithfulness score.
}
\end{table*}

\section{Examples of Generated Summaries}
\begin{tcolorbox}[colback=gray!5!white, colframe=gray!75!black, title=Data Sample 1, width=\linewidth]
\textbf{Source Text: }Their relationship led to jail time for her, but Mary Kay Letourneau Fualaau wants the world to know that she and her husband, Vili Fualaau, are still happily together. She was a married 34-year-old teacher and mother of four in Seattle in 1996 when she began an affair with Fualaau, her 13-year-old student. Letourneau gave birth to her young lover's child and went on to serve more than seven years in prison on charges related to their sexual relationship. On Friday, ABC's "20/20" will air an exclusive interview with the couple, who will celebrate their 10th wedding anniversary on May 20. The pair wed soon after she was released from prison in 2005 and are now the parents of two teen girls. According to ABC, during the interview with Barbara Walters, "Mary Kay tells Walters what makes their marriage work in spite of their huge difference in their age." "She also discusses her surprising plans to teach again, as well as her intentions to have her status as a registered sex offender lifted," a posting on the network's site says. "Vili Fualaau, meanwhile, discusses his bouts with alcoholism, depression and why he believes the system failed him while he was still a minor." Letourneau Fualaau is now 53, and her husband is 31.
\\
\textbf{Generated Summary:} Mary Kay Letourneau Fualaau, a married 34-year-old teacher and mother of four, was sentenced to over seven years in prison for having a relationship with her 13-year-old student, Vili Fualaau. The couple, now 53 and 31 respectively, are set to celebrate their 10th wedding anniversary. They are due to be interviewed by ABC's "20/20" on their marriage, teaching career, and plans to lift her status as a registered sex offender.
\end{tcolorbox}

\begin{tcolorbox}[colback=gray!5!white, colframe=gray!75!black, title=Data Sample 2, width=\linewidth]
\textbf{Source Text: }Kanye West has settled a lawsuit with a paparazzi photographer he assaulted -- and the two have shaken on it. The photographer, Daniel Ramos, had filed the civil suit against West after the hip-hop star attacked him and tried to wrestle his camera from him in July 2013 at Los Angeles International Airport. West pleaded no contest last year to a misdemeanor count of battery over the scuffle. A judge sentenced him to two years' probation, as well as anger management sessions and community service. Ramos and his lawyer, Gloria Allred, sought general and punitive damages in the civil suit, saying that West had interfered with the photographer's rights to pursue a lawful occupation. The case had been set for trial next week, but Allred issued a statement Tuesday night saying Ramos' side had filed a dismissal "because the case was settled to the satisfaction of the parties." She didn't disclose the details of the settlement other than saying that "one important aspect of it was an apology by Kanye West to our client, Daniel Ramos." Her statement included a picture of West and Ramos shaking hands, which she said happened after the apology. The original incident was caught on video, including the following exchange. "Kanye! Kanye! Talk to me, Kanye!" Ramos shouts outside a terminal at the Los Angeles airport on the night of July 19, 2013. "What's' going on? Why can't we talk to you? I mean, why?" he asks as West moves through a group of paparazzi. "Now come on, Kanye, I don't want to fight with you," he says as West advances toward him. "I told you, don't talk to me, right," West says. \"You're trying to get me in trouble so I step off and have to pay you like \$250,000.\" West is then seen rushing the photographer and attempting to wrestle his camera from his hands. West retreats after about 15 seconds of scuffling with the photographer. "We believe that this case sent an important message," Allred said. "Celebrities are not above the law, and they have no right to physically attack someone simply because they were asked a question." Beverly Hills Police investigated an incident in January 2014 in which West was accused of assaulting a man at a Beverly Hills chiropractor's office. West avoided criminal charges by reaching a civil settlement with the man. Kanye West apologizes to Beck, Bruno Mars. CNN's Sonya Hamasaki contributed to this report.
\\

\textbf{Generated Summary:} Kanye West has settled a lawsuit with a paparazzi photographer who accused him of assault at Los Angeles International Airport in 2013. The photographer had sought general and punitive damages, claiming West had interfered with his rights to pursue a lawful occupation. The case was set for trial next week, but was dismissed after a settlement was reached. West reportedly apologised to the photographer during the handshake, though the terms of the settlement were not disclosed. This follows earlier incidents where West settled civil lawsuits with photographers and a man at a Beverly Hills chiropractor's office.
\end{tcolorbox}

\end{document}